\documentclass{article}
\usepackage{iclr2020_conference,times}

\usepackage{amsmath,amsfonts,bm}

\def\eqref#1{equation~\ref{#1}}

\def\1{\bm{1}}

\DeclareMathAlphabet{\mathsfit}{\encodingdefault}{\sfdefault}{m}{sl}
\SetMathAlphabet{\mathsfit}{bold}{\encodingdefault}{\sfdefault}{bx}{n}

\usepackage{graphicx}
\usepackage{hyperref}
\usepackage{url}
\usepackage{capt-of}
\usepackage{comment}

\title{Fine-Grained classification for multi-source land cover mapping}

\author{Y. Jean Eudes Gbodjo, Dino Ienco \\
INRAE, UMR TETIS, University of Montpellier\\
Montpellier, France \\
\texttt{\{jean-eudes.gbodjo,dino.ienco\}@inrae.fr} \\
\And
Louise Leroux \\
CIRAD, UPR AÏDA \\
Dakar, Sénégal \\
\texttt{louise.leroux@cirad.fr} \\
\AND
Roberto Interdonato, Raffaele Gaetano \\
CIRAD, UMR TETIS \\
Montpellier, France \\
\texttt{\{roberto.interdonato,raffaele.gaetano\}@cirad.fr} \\
}

\iclrfinalcopy 

\newcommand{\method}{HOb2sRNN}

\begin{document}

\maketitle

\begin{abstract}
Nowadays, there is a general agreement on the need to better characterize agricultural monitoring systems in response to the global changes. Timely and accurate land use/land cover mapping can support this vision by providing useful information at fine scale. 
Here, a deep learning approach is proposed to deal with multi-source 
land cover mapping at object level. The approach is based on an extension of Recurrent Neural Network enriched via an attention mechanism dedicated to multi-temporal data context. Moreover, a new hierarchical pretraining strategy designed to exploit specific domain knowledge available under hierarchical relationships within land cover classes is introduced. Experiments carried out on the \textit{Reunion island} --a french overseas department-- demonstrate the significance of the proposal compared to remote sensing standard approaches for land cover mapping. 
\end{abstract}

\section{Introduction} \label{sec:intro}

Remote sensing has been used for decades to support agricultural monitoring systems that aim to provide up-to-date information, regarding food production, to stakeholders and decision makers~\citep{FRITZ2019258}. A typical application is the acreage estimation in cropland or crop type mapping that lies in the general field of Land use/Land Cover (LULC) mapping. Since, the agricultural sector is facing major challenges due to the global changes (climate, land competition, environmental pressures), there is an urgent need to better characterize agricultural monitoring systems at global and regional scales through timely and accurate information~\citep{Atzberger13}. 
Nowadays, a huge amount of satellite based remote sensing data is publicly available to improve the LULC characterization. In particular, the Sentinel-1 (S1) and Sentinel-2 (S2) missions are of interest since they provide at high spatial resolution (up to 10 meters) and high revisit time (up to 5 days), respectively, multi-temporal radar and optical images of continental areas. 

Despite the well know complementary of radar and optical sources~\citep{GaoMSH06, IannelliG18, IENCO201911}, their multi-temporal combination for LULC mapping is still a challenging task for which only few methods have been proposed~\citep{IencoGIOM19,IENCO201911}. However, promising results have recently paved the way for the multi-temporal radar and optical combination through deep learning techniques such as Convolutional Neural Networks (CNN) and Recurrent Neural Networks (RNN).  Furthermore, as regards LULC classes, specific knowledge can be derived. LULC classes can be categorized in a hierarchical representation where they are organized via class/subclass relationships. For instance, agricultural land cover can be organized in crop types and subsequently crop types in specific crops obtaining several levels of a taxonomy. As example, the Food and Agriculture Organization -- Land Cover Classification System~\citep{di2005land} is a kind of a hierarchical organization of LULC classes. Only, few studies~\citep{SULLAMENASHE2011392,Wu2016,SULLAMENASHE2019183} have considered the use of such hierarchical information which is, nonetheless, appealing for LULC mapping process. However, none of them have considered such kind of information in a multi-source fusion context.
This study aims to deal with the land cover mapping at object-level, using multi-source (radar and optical) and multi-temporal data, as well as specific domain knowledge about land cover classes. To this end, we propose a deep learning architecture, named \method{} (Hierarchical Object based two-Stream Recurrent Neural Network), which is based on an extension of RNN enriched via a customized attention mechanism capable to fit the specificity of multi-temporal data. In addition, a new strategy, named hierarchical pretraining, is introduced to get the most out of domain expert knowledge, available under hierarchical relationships between land cover classes. The study was conducted over the whole \textit{Reunion island} site where major economic issues around the sugarcane industry require timely and accurate monitoring of LULC mapping.
    
\section{Method} \label{sec:method}

\begin{minipage}[!htbp]{0.45\textwidth}
Figure~\ref{fig:method} depicts the proposed architecture for the multi-source and multi-temporal land cover mapping process. It consists of two branches, one for each source (radar or optical time series) which can be decomposed in two parts: i) the time series processing via a modified RNN we called FCGRU and ii) the multi-temporal combination of the FCGRU outputs through a modified attention mechanism to generate per-source features. Furthermore, the outputs of per branch FCGRU are concatenated and the modified attention mechanism is anew employed to generate fused features. Finally, the per-source and fused features are leveraged to predict the land cover classes. In addition, the architecture is trained exploiting specific knowledge about land cover classes represented under the shape of a hierarchy with class/subclass relationships. 
\end{minipage}
\hspace{0.5cm}
\begin{minipage}[!htbp]{0.5\textwidth}
\centering
\includegraphics[width=\columnwidth]{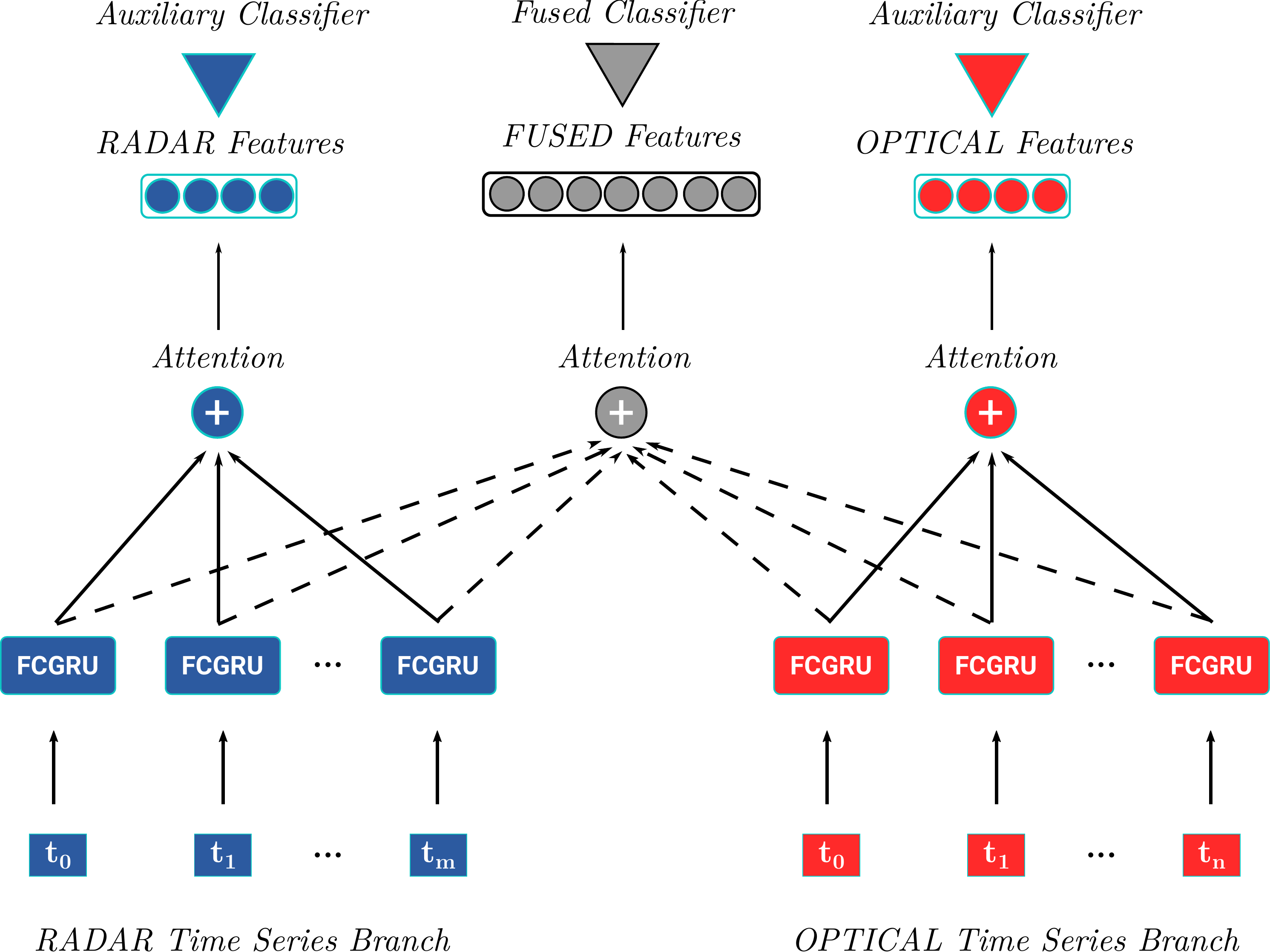}
\captionof{figure}{\label{fig:method}Visual representation of the HOb2sRNN method.}
\end{minipage}

Considering each branch, the first part is represented by a FCGRU cell that takes as input one sequence of the time series at each time stamp. The FCGRU cell is a modified structure of the standard GRU unit~\citep{ChoMGBBSB14}, a kind of RNN which has demonstrated its effectiveness in the field of remote sensing~\citep{Benedetti18,MouGZ17}. The FCGRU cell extend the GRU unit including two fully connected layers that process the input information, at a particular time stamp, before the use of the standard GRU unit. Such layers allow the architecture to extract an useful input combination for the classification task enriching the original data representation. A hyperbolic tangent ($tanh$) non linearity is associated to each of the layers for the sake of consistency, since the GRU unit is mainly based on Sigmoid and $tanh$ activations. 

As concerns the second part of the branches, a modified attention mechanism is employed on top of the FCGRU outputs (hidden states), yielded at each time stamp, to generate per-source features. Neural attention models~\citep{Bahdanau14,Luong15,BritzGL17} are widely used in 1D signal or natural language processing to combine the RNN outputs at different time stamps through a set of attention weights. In the standard attention model, the set of weights is computed using a $SoftMax$ function so that their values ranges in [0,1] and their sum is equal to 1. Due to this constraint, the $SoftMax$ attention has the property to prioritize one instance over the others making it well suited for tasks such as machine translation where each target word is aligned to one of the source word~\citep{KaramanolakisHG19}. However in the land cover mapping case, where multiple time stamps are generally useful to better distinguish among classes, relaxing the sum constraint could thus help to better weight the relevant time stamps, independently. Therefore, in our modified attention formulation, we substituted the $SoftMax$ function by a $tanh$ to compute weights. Apart from relaxing the sum constraint in $SoftMax$, the $tanh$ attention will return weights in a wider range i.e. [-1,1] also allowing negative values. 

The $tanh$ attention is also employed over the concatenation of the per branch FCGRU outputs to generate fused features. While the per-source features encode the temporal information related to the input source, fused features encode both temporal information and complementarity of radar and optical sources. Then, the set of features (per-source and fused) are leveraged to perform the land cover classification. The feature combination involves three classifiers: one classifier on top of the fused features ($feat_{fused}$) and two auxiliary classifiers, one for each source features ($feat_{rad}$ for radar and $feat_{opt}$ for optical). Auxiliary classifiers, as underlined by~\citet{INTERDONATO201991,IENCO201911}, are used to strengthen the complementarity and stress the discriminative power of the per-source features. The cost function associated to the optimization of the three classifiers is:
\begin{equation}
L_{total} = 0.5 \times L(feat_{rad}) + 0.5 \times L(feat_{opt}) + L(feat_{fused}) \label{eqn:cost}
\end{equation}
$L(feat)$ is the loss (categorical Cross-Entropy) associated to the classifier fed with the features $feat$. The loss of auxiliary classifiers was empirically weighted by 0.5 to enforce the discriminative power of the per-source features while privileging the fused features in the combination. The final land cover class is derived combining the three classifiers with the same weight schema employed in the cost function. 
As part of the exploitation of specific domain knowledge about a hierarchical organization of land cover classes, we learned the model following a specific scheme that considers each level of the taxonomy from the most general one (the most simple) to the most specialized (the target classification level) i.e. classification tasks of increasing complexity. Specifically, we start the model training on the highest level of the hierarchy and subsequently, we continue the training on the next level reusing the previous learned weights for the whole architecture, excepting the weights associated to the output layer (classifiers) since level-specific. This process is performed until we reach the target level. Such hierarchical pretraining strategy allows the model not only to focus first on easy classification problems and, gradually, to adapt its behaviour to deal with more complex tasks, but also to tackle the target classification level by integrating some kind of prior knowledge instead of addressing it completely from scratch. Code is available online~\footnote{\url{https://github.com/eudesyawog/HOb2sRNN}}.

\section{Data Description} \label{sec:data}
The study was carried out on the \textit{Reunion island}, a french overseas department located in the Indian Ocean, covering a little over 3000~$km^{2}$. The multi-source time series consists of 26 S1 images and 21 S2 images acquired between January and December 2017. S1 images were obtained from PEPS platform~\footnote{\url{https://peps.cnes.fr/}} at level-1C in C-band with dual polarization (VH and VV) and calibrated in backscatter values. A multi-temporal filtering was performed to reduce the speckle effect. S2 images were obtained from THEIA pole~\footnote{\url{http://theia.cnes.fr}} at level-2A in top of canopy reflectance. Only 10-m spatial resolution bands (Blue, Green, Red and Near Infrared) were considered for S2. Cloudy observations were linearly interpolated through a multi-temporal gapfilling~\citep{IngladaVATMR17}. As additional optical descriptor, we derived the NDVI index~\citep{Rouse1974} considered as a reliable indicator to distinguish among land cover classes especially vegetation. To ensure a precise spatial matching between images, all S1 images were orthorectified at the same 10-m spatial resolution and coregistered with the S2 grid. The ground truth of the \textit{Reunion island}~\footnote{Database is available online under \url{https://doi.org/10.18167/DVN1/TOARDN}} includes 6265 polygons distributed over 11 land cover classes (Table \ref{tab:gt}). In order to integrate specific knowledge in the land cover mapping process, we derive a taxonomic organization of the \textit{Reunion island} land cover classes (See Appendix \ref{app:hier}) obtaining two levels before the target classification level described in Table \ref{tab:gt}.

\begin{minipage}[!htbp]{0.4\textwidth}
As the level of analysis is object-based, a very high spatial resolution SPOT6/7 image was segmented obtaining 14\,465 segments. The ground truth was spatially intersected with the obtained segments to provide radiometrically homogeneous samples resulting in new 7\,908 polygons. Finally, each object was attributed with the mean value of the corresponding pixels over the multi-temporal and multi-source data, resulting in 157 variables per object ($26\times2$ for S1 $+$ $21\times5$ for S2).

\end{minipage}
\hspace{0.3cm}
\begin{minipage}[!htbp]{0.55\textwidth}
\centering
\captionof{table}{Characteristics of the Reunion island ground truth}
\label{tab:gt}
\scriptsize
 \begin{tabular}{cccc}
  \textit{Class} & \textit{Label} & \textit{Polygons} & \textit{Intersected} \\ \hline
  0 &  \emph{Sugarcane} & 869 & 1258 \\
  1 & \emph{Pasture and fodder} & 582 & 869 \\
  2 & \emph{Market gardening} & 758 & 912 \\
  3	& \emph{Greenhouse crops} & 260 & 233 \\
  4	& \emph{Orchards} & 767 & 1014 \\
  5	& \emph{Wooded areas} & 570 & 1106 \\
  6	& \emph{Moor and Savannah} & 506 & 850 \\
  7	& \emph{Rocks and bare soil} & 299 & 573 \\
  8	& \emph{Relief shadows} & 81 & 107 \\
  9 & \emph{Water} & 177 & 261 \\
 10	& \emph{Urbanized areas} & 1396 & 725 \\ 
 Total & & 6265 & 7908 \\ 
 \hline
 \end{tabular}
\end{minipage}

\section{Experimental Evaluation} \label{sec:exp}
In this section, we present and discuss the results obtained on the study site introduced in Section~\ref{sec:data}. To evaluate the behavior of \method, we chosen several baselines i.e. Random Forest (RF) and Support Vector Machine (SVM) which are commonly employed to deal with land cover mapping~\citep{Erinjery2018345} and a Multi Layer Perceptron (MLP). The baselines were trained on the concatenation of the multi-temporal radar and optical data. To learn the \method{} and MLP parameters, we used Adam optimizer~\citep{KingmaB14} with a learning rate of $1 \times 10^{-4}$, and conducted training over 2000 epochs with a batch size of 32. Considering the \method{} model, the number of hidden units of the FCGRU cell was fixed to 512 for each branch and 64 then 128 neurons were employed for the fully connected layers respectively. Concerning the MLP competitor, two hidden layers and the same number of hidden units i.e. 512 were employed. Regarding other competitors, RF was optimized via the maximum depth, the number of trees and the number of features while SVM was optimized via the kernel type, the gamma and the penalty terms. The dataset were split into training, validation and test set with an object proportion of 50\%, 20\% and 30\% respectively. The values were normalized per band (resp. indices) in the interval [0,1]. Training set was used to learn the models while validation set was exploited for model selection. Finally, the model achieving the best accuracy on the validation set was employed to classify the test set. The assessment of the classification performances was done considering \textit{Accuracy}, \textit{F1 Score} and \textit{Kappa} metrics. Results were averaged over ten random splits since performances may vary depending on the data split.

\begin{minipage}[!htbp]{0.6\textwidth}
\captionof{table}{F1 score, Kappa and Accuracy considering the different methods}
\label{tab:avg}
\begin{center}
\scriptsize
\begin{tabular}{c c c c} 
& \textit{F1 Score} & \textit{Kappa} &\textit{Accuracy} \\ \hline
RF & 75.62 $\pm$ 1.00 & 0.726 $\pm$ 0.011 & 75.75 $\pm$ 0.98 \\ 
SVM & 75.34 $\pm$ 0.88 & 0.722 $\pm$ 0.010 & 75.39 $\pm$ 0.89 \\
MLP & 77.96 $\pm$ 0.70 & 0.752 $\pm$ 0.008 & 78.03 $\pm$ 0.66 \\ 
HOb2sRNN & \textbf{79.66} $\pm$ \textbf{0.85} & \textbf{0.772} $\pm$ \textbf{0.009} & \textbf{79.78} $\pm$ \textbf{0.82} \\ \hline
\end{tabular}
\end{center}

\captionof{table}{F1 score, Kappa and Accuracy considering different ablations of HOb2sRNN}
\label{tab:ablation}
\begin{center}
\scriptsize
 \begin{tabular}{c c c c}
& \textit{F1 Score} & \textit{Kappa} &\textit{Accuracy} \\ \hline 
noEnrich & 79.09 $\pm$ 0.57 & 0.764 $\pm$ 0.006 & 79.10 $\pm$ 0.50 \\
noHierPre & 78.35 $\pm$ 0.70 & 0.756 $\pm$ 0.007 & 78.43 $\pm$ 0.66 \\
noAtt & 77.66 $\pm$ 0.99 & 0.749 $\pm$ 0.011 & 77.74 $\pm$ 0.99 \\
SoftMaxAtt & 77.32 $\pm$ 1.22 & 0.746 $\pm$ 0.013 & 77.47 $\pm$ 1.18 \\
HOb2sRNN & \textbf{79.66} $\pm$ \textbf{0.85} & \textbf{0.772} $\pm$ \textbf{0.009} & \textbf{79.78} $\pm$ \textbf{0.82} \\ \hline
\end{tabular}
\end{center}
\end{minipage}
\hspace{0.5cm}
\begin{minipage}[!htbp]{0.36\textwidth}
Table \ref{tab:avg} reports the average results of the different methods. We can observe considering the average behavior that \method{} outperformed its competitors gaining about 2 points with respect to the best competitor i.e. MLP. The MLP competitor which is a less explored method for land cover mapping achieved better performances than the common RF and SVM algorithms. In Table~\ref{tab:ablation} we investigated the interplay between the main components of the \method{} architecture and we disentangled their benefits. 
\end{minipage}
\vspace{0.1cm}

We excluded firstly the enrichment step in the FCGRU cell naming this variant \textit{NoEnrich}, then the hierarchical pretraining process naming \textit{NoHierPre} and finally the three attention mechanisms involved in the architecture naming \textit{NoAtt}.
 We also investigated another variant naming \textit{SoftMaxAtt} in which we replace, in the \method{} architecture, our $\tanh$ based attention mechanism with standard $SoftMax$. We can first note the benefit of the modified attention mechanism obtaining about 2 points more than the NoAtt and SoftMaxAtt variants which perform similar. This behavior confirms our hypothesis that relaxing the sum constraint is more beneficial for multi-temporal land cover classification. As regards the hierarchical pretraining, we can also note the added value of such step gaining about 1 point. It seems to underline that including specific knowledge in the pretraining process of neural networks can improve final classification performances. Lastly, the enrichment step in the FCGRU cell also proved a certain contribution to the final results.

\section{Conclusion}
In this work, we dealt with land cover mapping at object level using multi-source and multi-temporal data, as well as specific domain knowledge about land cover classes. To this end, we designed a deep learning architecture named \method{}, especially tailored to leverage sources complementarity and dependencies carried out by multi-temporal data, through specific branches and dedicated attention mechanism. In addition, the architecture was coupled with a new pretraining strategy, as part of the exploitation of domain expert knowledge associated to a hierarchical organization of land cover classes. The proposal outperformed standard approaches to deal with LULC mapping.

\bibliography{references}
\bibliographystyle{iclr2020_conference}

\appendix
\section{Taxonomic organization of the Reunion island land cover classes} \label{app:hier}

\begin{figure}[!htbp]
\centering
\includegraphics[width=.89\columnwidth]{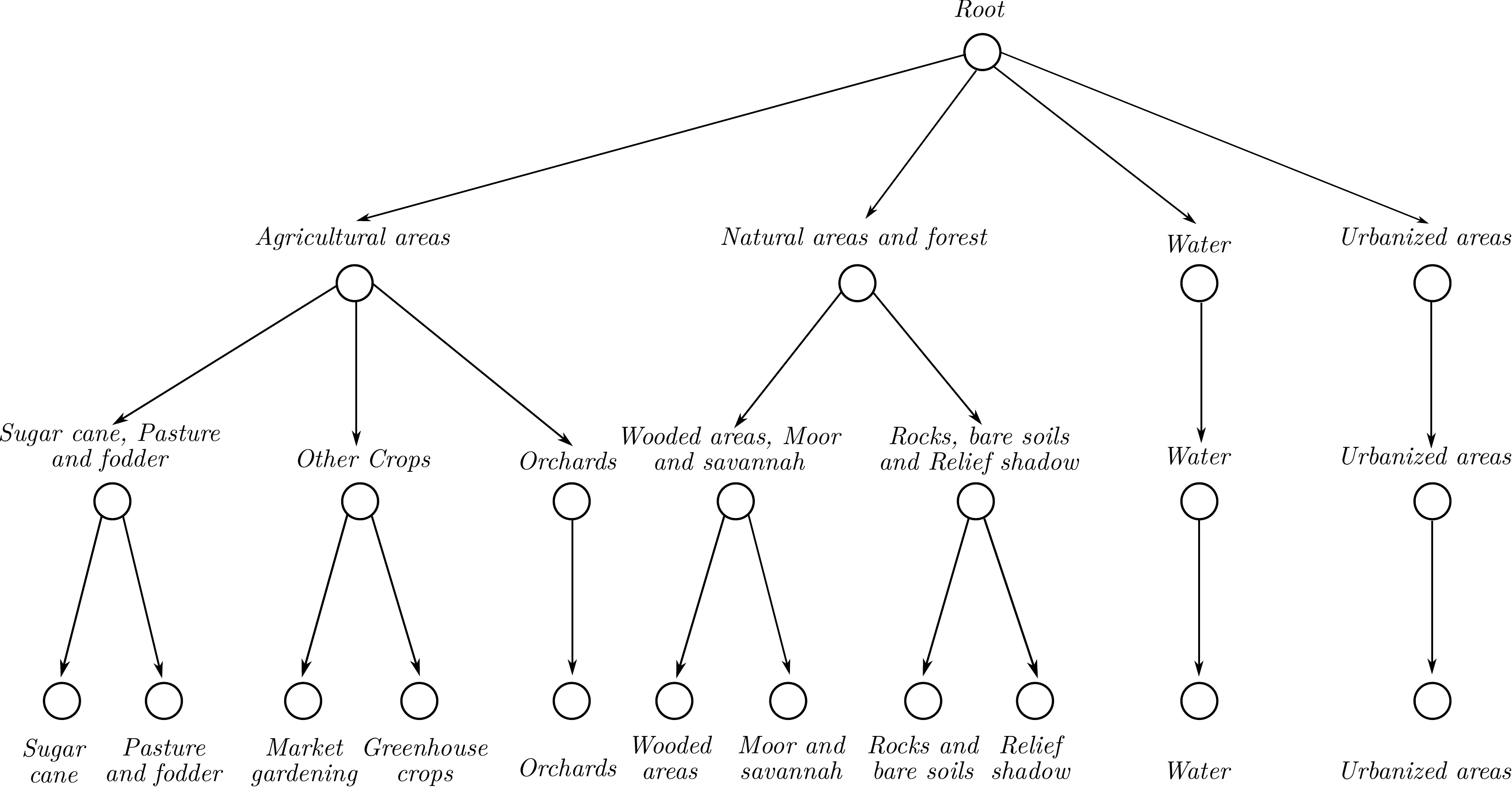}
\end{figure}

\end{document}